\documentclass[10pt,a4paper]{article}
\usepackage[utf8]{inputenc}
\usepackage{amsmath}
\usepackage{amsfonts}
\usepackage{amssymb}
\usepackage{url}
\usepackage{authblk}
\usepackage{graphicx}

\newcommand{\nome}{IAT{ }}
\newcommand{\nomeurl}{\url{http://www.ivl.disco.unimib.it/research/imgann}{ }}
\newcommand {\quot}[1]{``#1''}

\begin{document}
\title{\nome - Image Annotation Tool: Manual}
\author{Gianluigi Ciocca, Paolo Napoletano, Raimondo Schettini}
\affil{DISCo - Department of Informatic, Systems and Communication \\ University of Milano-Bicocca \\ Viale Sarca 336, 20126, Milano, Italy}
\date{February 2015}
\maketitle

\section{Introduction}
The annotation of image and video data of large datasets is a fundamental task in multimedia information retrieval \cite{Datta06,Hu2011} and computer vision applications \cite{Everingham10,Maybank2004,Wang2013}. In order to support the users during the image and video annotation process, several software tools have been developed to provide them with a graphical environment which helps drawing object contours, handling tracking information and specifying object metadata \cite{halaschek2005photostuff,labelme,labelmevideo,vatic,viper,flowboost,bianco2014interactive-tool}. For a survey on image and video annotation tools interested readers can refer to \cite{Dasiopoulou11}. Here we introduce a preliminary version of the image annotation tools developed at the Imaging and Vision Laboratory.

\section{The application}

\nome is an application for the annotation of digital images. It allows the users to annotate regions on the images in several ways (predefined or free), and to associate to the selected regions labels depending on the application context chosen from a predefined taxonomy. The application allows to choose whether annotate a single image, or start a project to annotate several images in a sequential manner. The results are saved and stored in files in textual format. It is possible for the user to load the results and resume the annotation process from the last saved state. The annotations are editable either by moving the reference points that characterize them allowing the user to annotate in the best way the portion of the image. Three different shapes of annotation are provided: rectangular, elliptical and polygonal. These three shapes are able to cover a large range of region shapes. 

To facilitate the  annotation process, the labels to be associated to the annotated regions are organized in a three-level taxonomy: Class, Type and Name. The Class is used to groups entities that share similar macro properties (e.g. vehicles, foods, people). Within a class we can have entities of different types (e.g. car, bicycle, vegetable, fruit, male, female). Finally the name is a unique identifier of an entity present in the current image being annotated. Class and Type are predefined and depend on the application domain in which the image annotation are used. 

The application has been built with the requirement of being cross-platform in mind. For this reason the Qt environment and C++ languages were used. Qt \cite{qt} is an environment for open source software development. It is based on multi-platform libraries, and for this reason Qt is used when applications using graphical interfaces are needed. 

The following images illustrate an example of the application user interface and the image annotation process.

\begin{figure}
\center
\includegraphics[scale=0.35]{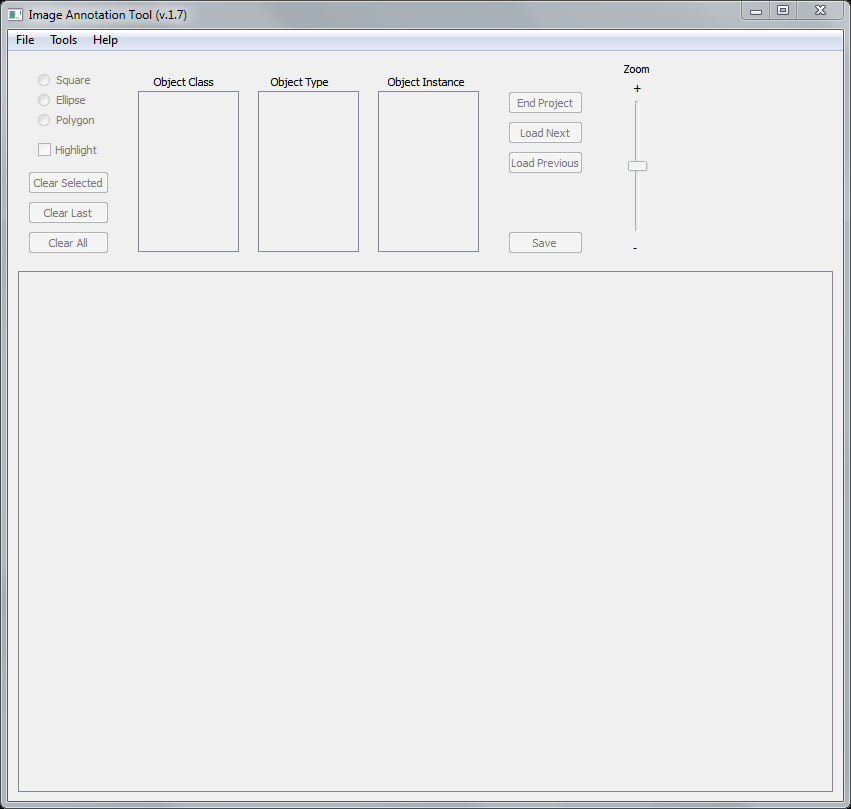}
\label{fig:image1}
\caption{The GUI of the program. At the top, the annotation tools. At the bottom, the image window.}
\end{figure}

\begin{figure}
\center
\includegraphics[scale=0.35]{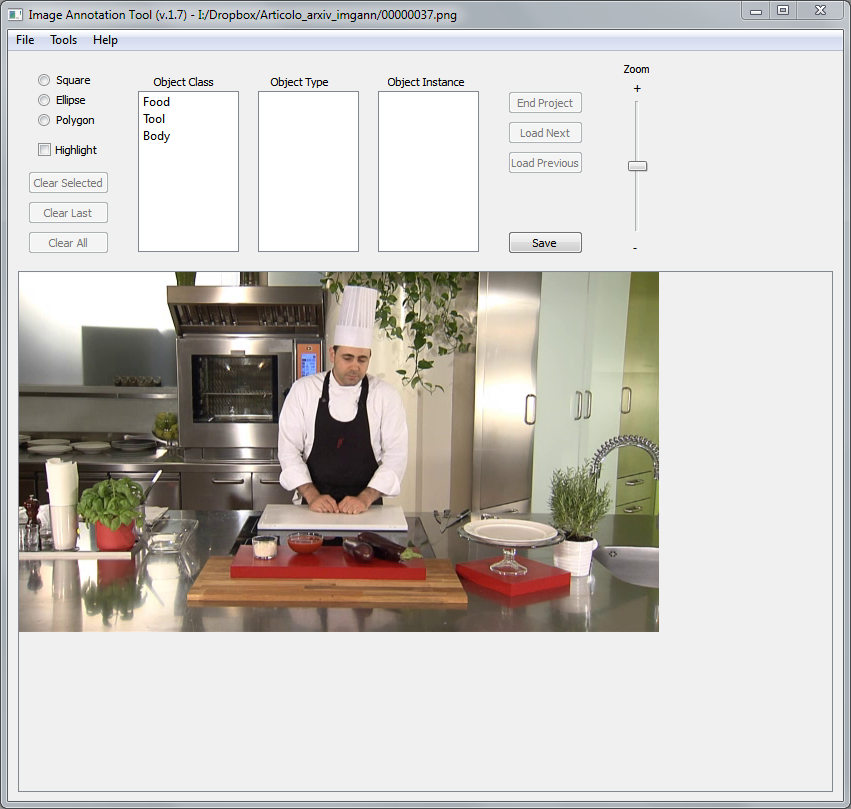}
\label{fig:image2}
\caption{Selection of the \quot{Annotate a Single Image}. An image is selected and shown at the bottom of the window. After loading the image, a file containing the labels to be used is also selected and loaded. At the top, the first level of the taxonomy is shown in the \quot{Object Class} list.}
\end{figure}

\begin{figure}
\center
\includegraphics[scale=0.35]{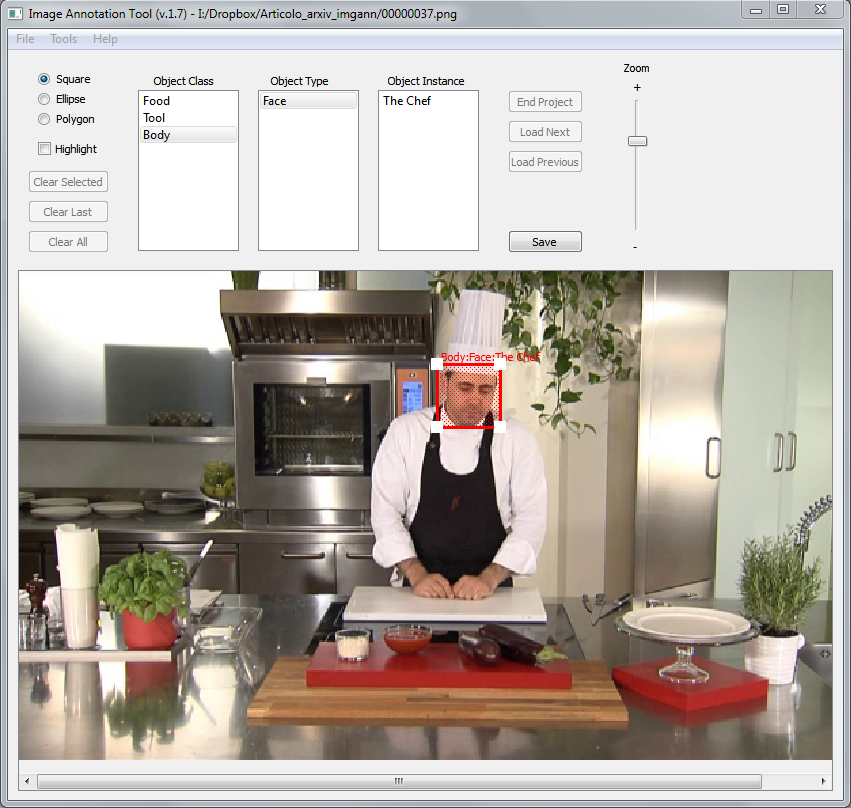}
\label{fig:image3}
\caption{The image has been zoomed in and the face has been annotated. At the top it can be seen the selected hierarchy of the label.}
\end{figure}

\begin{figure}
\center
\includegraphics[scale=0.35]{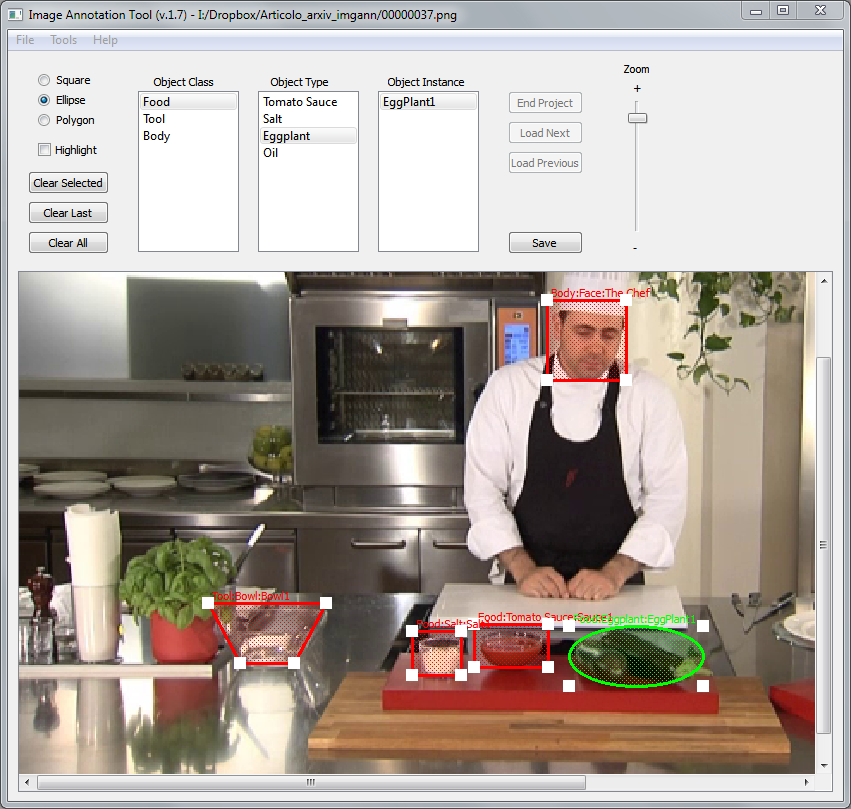}
\label{fig:image4}
\caption{Several objects have been selected using the different shapes at disposal. The currently selected shape is shown in green.}
\end{figure}

\begin{figure}
\center
\includegraphics[scale=0.35]{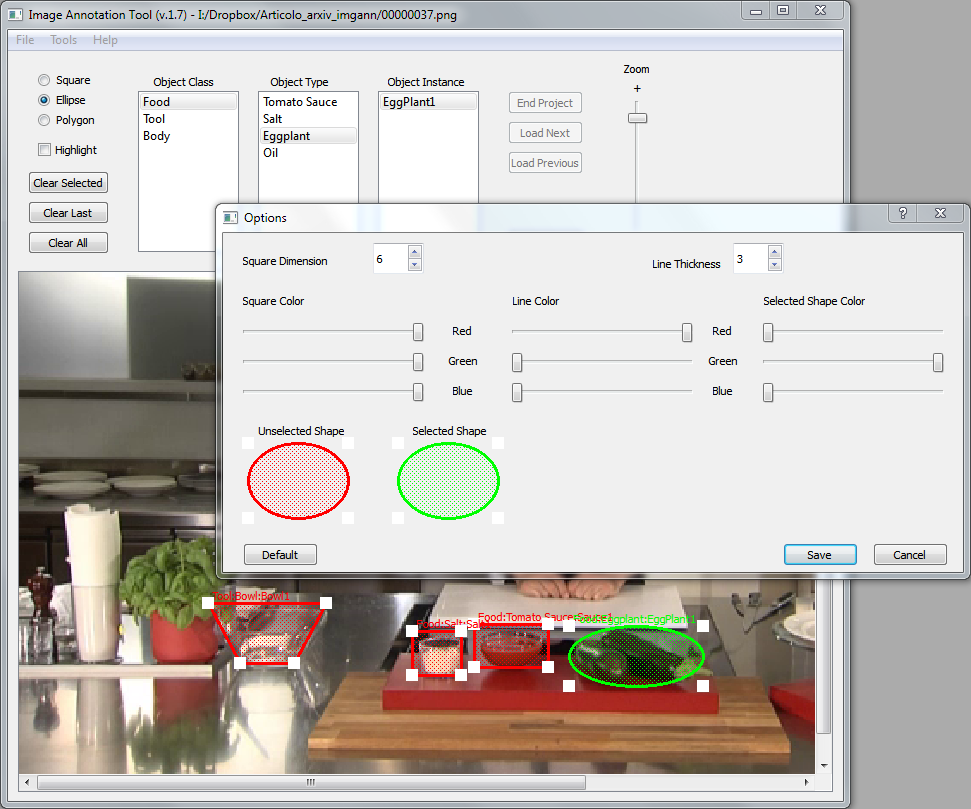}
\label{fig:image5}
\caption{Some properties of the displayed annotation can be changed from the \quot{Option} menu command.}
\end{figure}

\begin{figure}
\center
\includegraphics[scale=0.35]{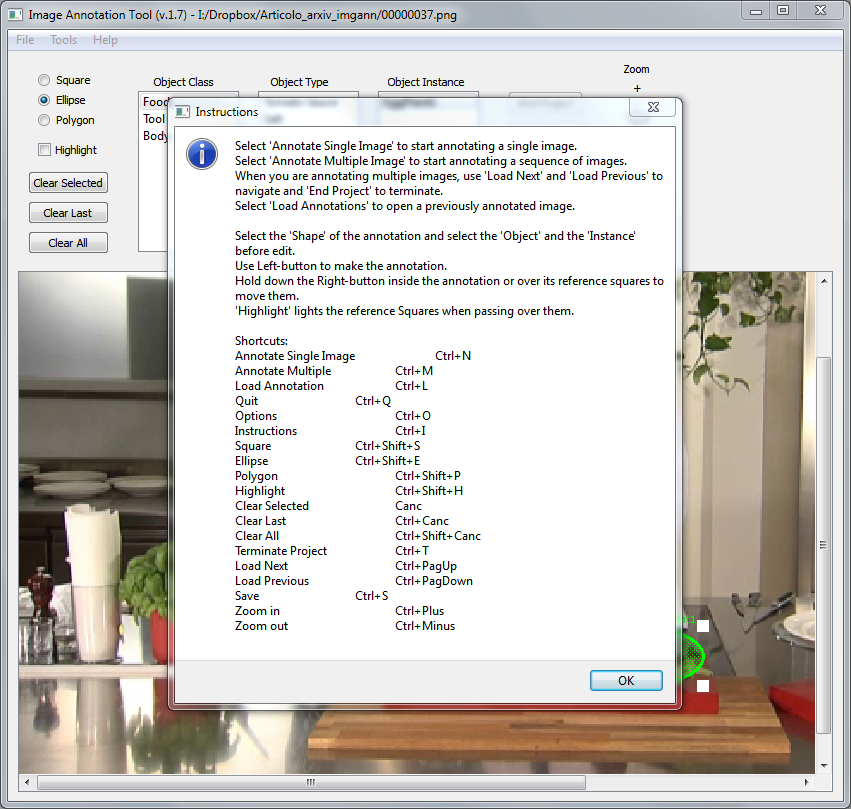}
\label{fig:image6}
\caption{The \quot{Help} menu command shows some basic instructions on the use of the application and the list of keyboard shortcuts.}
\end{figure}

\section{Usability}

A usability test phase has been performed at the end of program development. The objective was to assure the quality of the GUI, evaluating if the program properly reacts to the inputs and user's needs. 

To this end, several tester have been engaged in the evaluation phase. They were charged to perform several tasks o increasing difficulty to cover all the aspects of the application such as annotate an image, load an image annotation file, edit or delete some annotations, create polygonal annotations, start a new project.

None of the chosen testers have found difficulties to complete the simple tasks. There have been slight difficulties in creating polygonal annotations. Testers with little and medium experience in using computer, found difficulties to start a new project. Expert users found all the aspects pretty simple and practical despite some initial doubts.

At the end of the test, users were given a short questionnaire to evaluate some aspects of the program using a 1 (one) to 5 (five) scale. They were given 14 statements and asked if they strongly disagree (1), disagree (2), are neutral (3), agree (4), or strongly agree (5).

Table \ref{tab:feedbacks} shows the statements and the average user feedback for each statement. It should be noted that some results must be considered positive for low feedback values (e.g. 1, 4, 6, 8, 10, and 13) while other for high feedback values (2, 3, 5, 7, 9, 11, 12, and 14).

\begin{table}
\label{tab:feedbacks}
\small
\caption{Average user feedback on the 14 statements. 1$\rightarrow$Strongly Disagree, 2$\rightarrow$Disagree, 3$\rightarrow$Neutral, 4$\rightarrow$Agree, 5$\rightarrow$Strongly Agree.}
\begin{tabular}{llc}
\hline
\# & Statement	& Feedback \\ \hline	   
1 & I found the system unnecessarily complex sometime	& 2.28 \\	   
2 & I thought the system was easy to use & 4.14 \\
3 & I found the various functions in this system were well integrated	& 4.00 \\	   
4 & I found the system very cumbersome to use	& 1.71	 \\   
5 & I imagine that people would learn to use the system quickly	& 3.71 \\	   
6 & I found the instruction window a bit incomplete	&3.14	 \\   
7 & I found the features intuitive	& 4.28	\\   
8 & The project phase should be improved &	3.28	\\   
9 & The keyboard short-cuts are useful	& 3.71	   \\
10 & It is difficult to keep track of the already annotated items	& 1.85	\\   
11 & The annotation can be drawn quickly	& 4.28	  \\ 
12 & The system user interface is easy to understand	& 3.71	\\   
13 & The program seems to lack in functionality	& 2.71\\	   
14 & Shortcuts were intuitive	& 3.85	\\   
\hline
\end{tabular}
\end{table}

On the overall, the system proved to be intuitive considering that all the users, except one, were able to accomplish all the goals proposed. The only failure was due to a program crash during the achievement of a single goal. The testers also provided valuable suggestions. For example, some users asked to provide the possibility to create and edit the files that contain the taxonomy of labels to be used during annotation.

\section{Conclusion}
\nome is our first step toward a more complete image annotation tool. Its initial functionality and graphical user interface have been demonstrated to be efficient and effective. Our aims is to add more functionality to make the annotation process simpler and faster. For example we plan to add advanced image processing algorithms to help the user to semi-automatically annotate regions of interest.

\nome can be freely downloaded from \nomeurl . Also, at the same page, a video demo illustrating the image annotation process is provided.

\end{document}